\def\notes{0}

\documentclass[11pt]{article}

\usepackage{fullpage}
\usepackage{url}
\usepackage[usenames,dvipsnames,svgnames,table]{xcolor}
\usepackage{amssymb,amsthm,amsfonts,amsmath}
\usepackage{wrapfig}
\usepackage{graphicx}
\usepackage{subcaption}
\usepackage{multirow}
\usepackage{xspace}
\usepackage{paralist}
\usepackage[numbers,longnamesfirst]{natbib}
\usepackage{hyperref}

\definecolor{darkgray}{rgb}{.5,.5,.5}
\definecolor{forest}{rgb}{.5,.8,.5}
\newcommand{\remove}[1]{}

\newtheorem{definition}{Definition}
\newtheorem{theorem}{Theorem}

\newtheorem{prop}[theorem]{Proposition}
\newtheorem{lemma}[theorem]{Lemma}

\theoremstyle{definition}

\newcommand{\cA}{{\mathcal{A}}}
\newcommand{\cH}{{\mathcal{H}}}

\newcommand{\cX}{{\mathcal{X}}}
\newcommand{\cP}{{\mathcal{P}}}

\newcommand{\abs}[1]{\left| {#1}\right|}
\newcommand{\err}{\text{\rm err}}

\newcommand{\bx}{{\mathbf{x}}}
\newcommand{\bX}{{\mathbf{X}}}

\DeclareMathOperator*{\Exp}{{\mathbb{E}}}

\newcommand{\eps}{\epsilon}

\newcommand{\anl}{A}
\newcommand{\sig}{\gamma}
\newcommand{\mech}{M}
\newcommand{\score}{\text{score}}

\newcommand{\paren}[1]{{\left( {#1} \right)}}

\newcommand{\bparen}[1]{{\big( {#1} \big)}}

\ifnum\notes=1
\newcommand{\as}[1]{\marginpar{\tiny\sf {#1}}}
\else
\newcommand{\as}[1]{}
\fi

\begin{document}
\title{Information, Privacy and Stability \\ in Adaptive Data Analysis}
\author{Adam Smith\thanks{Computer Science and Engineering Department, 
Pennsylvania State University, University Park, PA,
USA. \texttt{asmith@psu.edu}. Supported by NSF award IIS-1447700, a Google Faculty
Award and a Sloan Foundation research award.}
}
\date{February 2017}

\maketitle

\begin{abstract}
  Traditional statistical theory assumes that the analysis to be
  performed on a given data set is selected independently of the data
  themselves. This assumption breaks downs when data are re-used
  across analyses and the analysis to be performed at a given stage
  depends on the results of earlier stages. Such dependency can arise when the same
  data are used by several scientific studies, or when a single
  analysis consists of multiple stages.

 How can we draw statistically valid
  conclusions when data are re-used? This is the focus of a recent and
  active line of work. At a high level, these
  results show that limiting the
  \emph{information} revealed by earlier stages of analysis controls
  the \emph{bias} introduced in later stages by adaptivity. 

  Here we review some known results in this area and highlight the
  role of information-theoretic concepts, notably several
  one-shot notions of mutual information.
\end{abstract}

\tableofcontents

\newpage
\section{Introduction}
\label{sec:intro}

How can one do meaningful statistical inference and machine learning
when data are \emph{re-used} across analyses? The situation is common
in empirical science, especially as data sets get bigger and more complex.
For example, analysts often clean the data and perform various
exploratory analyses---visualizations, computing descriptive
statistics---before selecting how data will be treated. Many times
the main analysis also proceeds in stages, with some sort of feature
selection followed by inference using the selected features. In such
settings, the analyses performed in later stages are chosen
\emph{adaptively} based on
the results of earlier stages that used the same data. Adaptivity
comes into even sharper relief when data are shared across multiple studies, and
the choice of the research question in subsequent studies may depend
on the outcomes of earlier ones. 
Adaptivity has been singled out as
the cause of a ``statistical crisis'' in science~\cite{GelmanL13}.

There is a large body of work in statistics and machine learning on
preventing false discovery, for example by accounting for multiple
hypothesis testing. Classical theory, however, 
assumes that the analysis is fixed independently of the
data---it breaks down completely when analyses are selected
adaptively. Natural techniques, such as separating a
validation set (``holdout'') from the main data set to verify
conclusions or the bootstrap method, do not
circumvent the issue of adaptivity: once the holdout has been used,
any further hypotheses tested using the same holdout will again depend
on earlier results. \citet{BH15} point out that this issue arises with leaderboards for machine learning
competitions: they observe that one can do well on the leaderboard
simply by using the feedback provided by the leaderboard itself on an
adaptively selected sequence of submissions---that
is, without even consulting the training data!

\begin{figure}
  \centering
   \includegraphics[height = 2in]{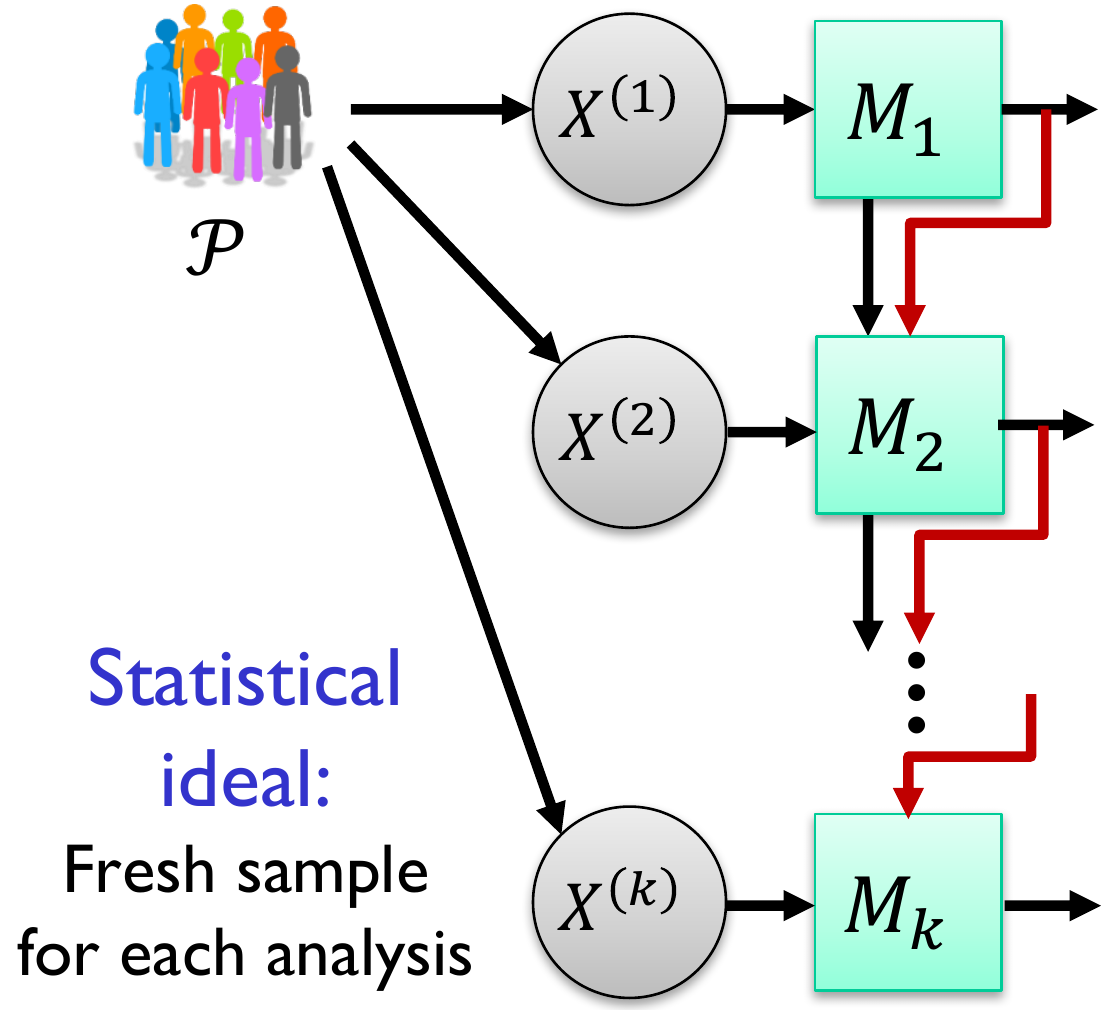} \hspace{1in}
   \includegraphics[height = 2in]{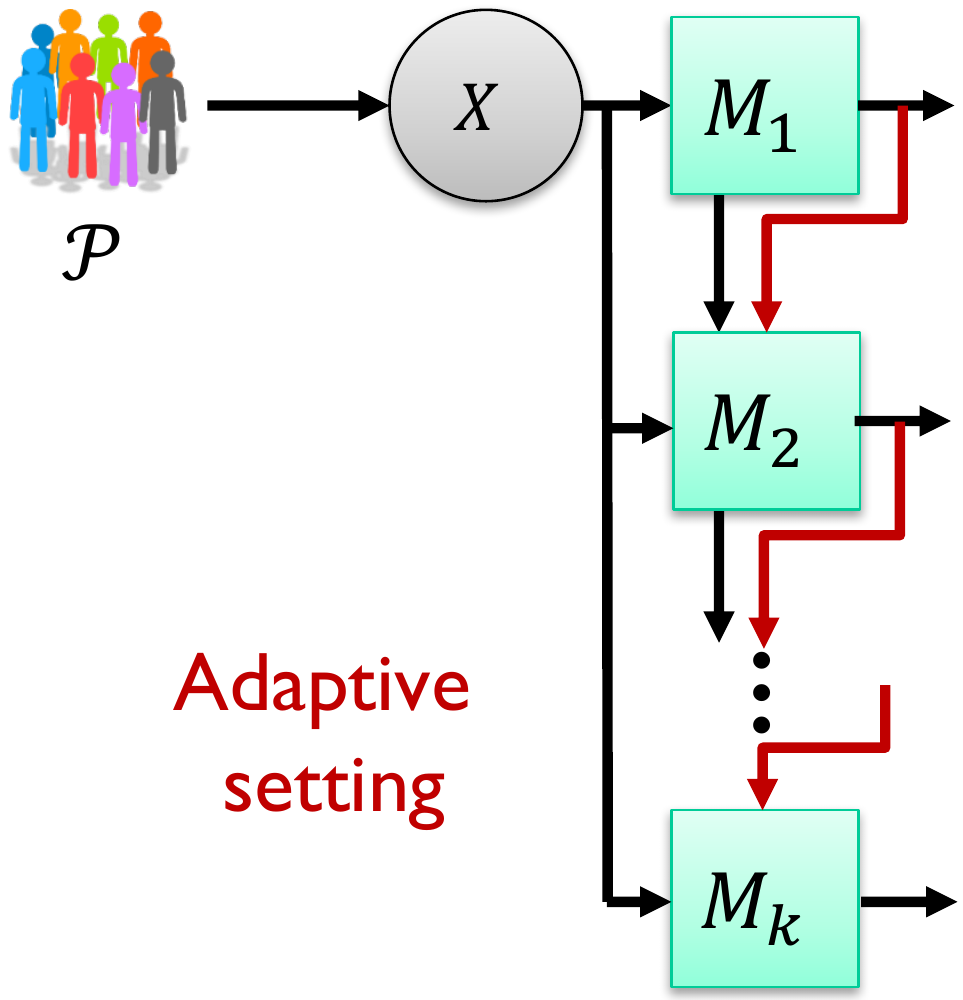}
\caption{\footnotesize Ideally, we would collecty fresh data from the same
  population $\cP$ for each analysis (left). In many real settings, we have only a
  single data set that must be re-used, leading to adaptively
  selecected analyses (right).  The arrows pointing into the top of
  the analyses indicate that each  analysis is selected based on the results of previous stages. }
  \label{fig:real-v-ideal}
\end{figure}

To formalize our situation somewhat, imagine there is a population that
we wish to study, modeled by a probability distribution $\cP$. An
analyst selects a sequence of  analyses
$\mech_1, \mech_2, ...,$ that she wishes to perform (we specify the type
of analysis to consider later). In an ideal world (Figure
\ref{fig:real-v-ideal}, left), the analyst would run each analysis $\mech_i$
on a fresh sample $\bX^{(i)}$ from the population. For simplicity,
we only discuss i.i.d samples in this article; we may assume that each
sample $\bX^{(i)}$ has $n$ points,  drawn independently from $\cP$. In the real
(adaptive) setting, the same data set $\bX$ gets used for each
analysis (Figure~\ref{fig:real-v-ideal}, right). The challenge is that
we ultimately want to learn about $\cP$, not $\bX$, but adaptive
queries can quickly overfit to $\bX$.

Our goal is to relate these two settings---to develop techniques
that allow us to emulate the ideal world in the real one, and
understand how much  accuracy is lost due to adaptivity. As mentioned
above, merely setting aside a holdout to verify results at each stage
is not sufficient, since the holdout ends up being re-used
adaptively. If the number $k$ of analyses is known ahead of time, one
can split the data into $k$ pieces of  $n/k$ points each
(assuming i.i.d. data, the pieces are  independent). This practice,
called \emph{data splitting}, provides clear validity guarantees, but
is inefficient in its use of data: data splitting requires $n$ to be
substantially larger than $k$, while we will see techniques that do
substantially better. Data splitting also requires an agreed upon
partition of the data, which can be problematic with data shared
across studies.

A
line of work in computer science
\cite{DFHPRR15,HardtU14,DFHPRR15science,DFHPRR15nips,russo2015controlling,SteinkeU15,BNSSSU16,RRST16,WangLF16,Elder16a,Elder16b}
initiated by \citet*{DFHPRR15} and \citet*{HardtU14} provides a set of tools and specific
methodology for this problem. This article briefly surveys the ideas
in these works, with emphasis on the role of several
information-theoretic concepts. Broadly, there is a strong connection
between the extent to which  an adaptive sequence of analyses remains
faithful to the underlying population $\cP$, and the amount of
information that is leaked to the analyst about $\bX$. In particular,
randomization plays a key role in the state of the art methods, with
a notion of algorithmic stability---\emph{differential
  privacy}---playing a central role.\as{Emphasize freedom to select analyses.}

Another approach, with roots in the statistics community, seeks to model particular
sequences of analyses, designing methodology to adjust for the bias
due to conditioning on earlier results (e.g., ~\cite{potscher1991effects,hurvich1990impact,fithian2014optimal,efron2014estimation,lockhart2014significance,lee2016exact}). The
specificity of this line of work makes it hard to compare with the
 more general approaches from
computer science. Other work in statistics hews an intermediate path,
allowing the analyst freedom within  a prespecified class of
analyses~\citep{berk2013valid,buja2015models}. There are intriguing
similarities between these lines of work and the work surveyed here,
such as the use of randomization to break up dependencies (e.g., \citep{TianT15,TianBT16,HarrisPMBT16});
understanding these connections more deeply is an important direction
for future work.

\section{The Lessons of Linear Queries}
\label{sec:SQ}

A simple but important setting for thinking about adaptivity,
introduced by \citet{DFHPRR15} and \citet{HardtU14}, is that
of an analyst posing an adaptively selected sequence of queries, each
of which asks for the expectation of a bounded function in the
population. Such queries capture a wide range of basic descriptive statistics
(the prevalence of a disease in a population, for example, or the
average age). Many inference algorithms can also be expressed in terms
of a sequence of such queries~\cite{Kea98}; for example, optimization
algorithms that query the gradient of a Lipschitz, decomposable loss function.

Suppose each data point lies in a universe $\cX$, so that a data set
lies in  $\cX^n$ and the underlying population is a distribution on
$\cX$. 
A \emph{bounded linear query} is specified by a function
$\phi:\cX\rightarrow [0,1]$.   The \emph{population value} of a linear
query is simply the expected value of the function when evaluated on
an element of the data universe drawn according to $\mathcal{P}$,
denoted $\phi(\cP) = \mathbb{E}_{x \sim \cP}[\phi(x)]$.  \footnote{The
``linear'' in ``bounded linear query'' refers to the fact that we care
about the expectation of  a function, so the resulting functional is
a linear map from the set of distributions om $\cX$ to $[0,1]$. In contrast,
some statistics, such as the variance of a random variable, are not
linear. ``Vounded'' refers to the image being limited to
$[0,1]$ (or, equivalently, any other finite interval). }


\begin{figure}
  \centering
  \includegraphics[height = 2in]{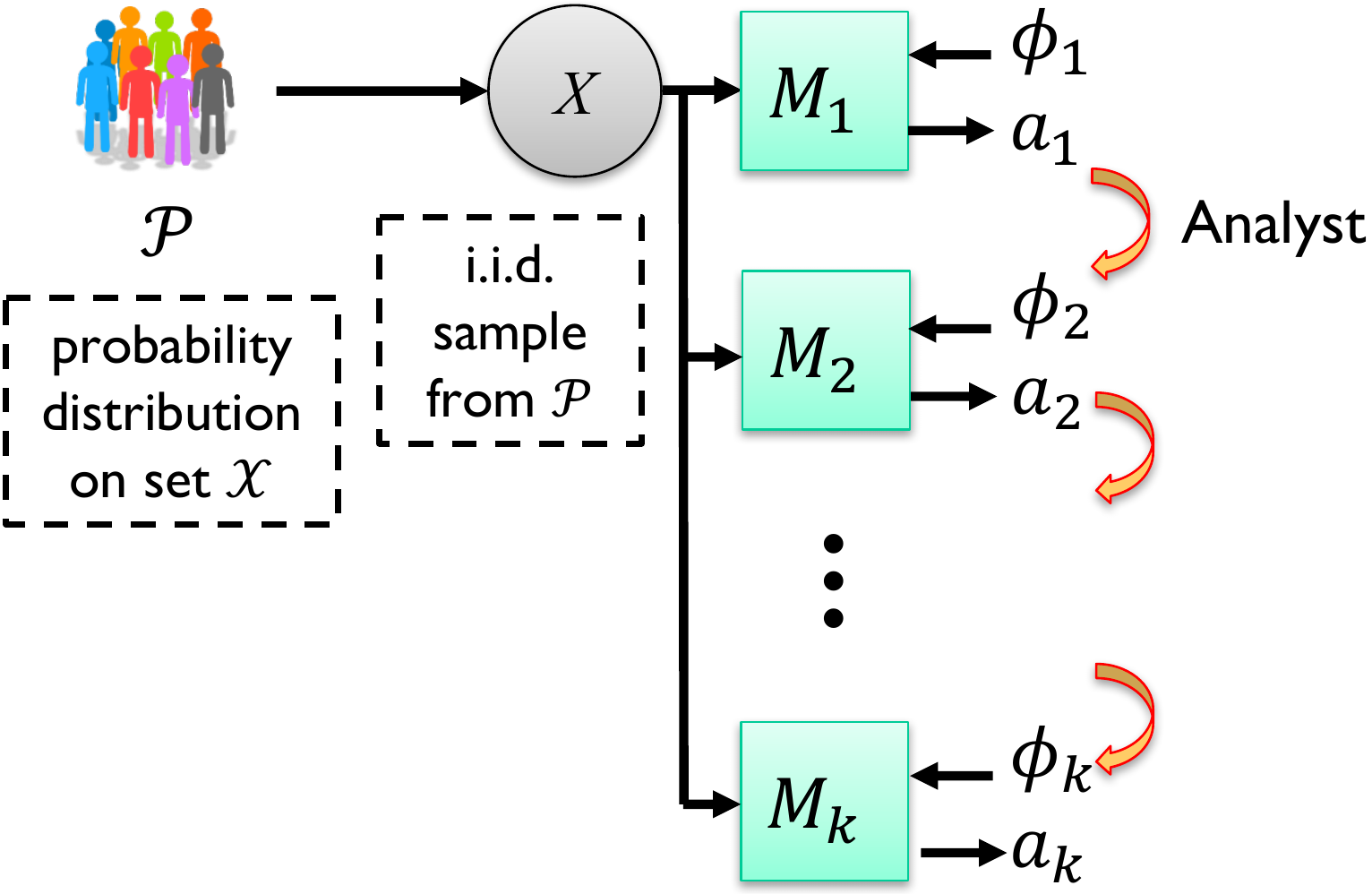}
  \caption{Adaptively selected linear queries}
  \label{fig:SQ}
\end{figure}

Consider now an interaction between an analyst wishing to pose such
queries and an algorithm $\mech$ (called the mechanism)
\as{image
  uses $M_1,...,M_k$.}%
holding a data
set $\bX$ sample i.i.d from $\cP$ that attempts
to provide
approximate answers to the queries $\phi_1, \phi_2,...$. This is
illustrated in
Figure~\ref{fig:SQ}, where we use subscripts (as in $M_1,M_2,...$) to
distinguish $k$ different rounds of $M$.  In general, neither the mechanism
nor the analyst knows the exact distribution $\cP$ (otherwise, why
collect data?), so the mechanism cannot always answer $\phi(\cP)$. A
natural approach is to answer with the \emph{empirical mean}
$\phi(\bX)=\frac{1}{n}\sum_{x_i \in \bx}\phi(X_i)$. When queries are
selected nonadaptively, this is the best estimator of $\phi(\cP)$.
We shall see,
however, this is not the best mechanism for estimating the
expectations of adaptively selected queries!

Given a query answering mechanism $\mech$, a data analyst $\anl$, and
a distribution $\cP$ on the data universe $\cX$, consider a random
interaction defined by selecting a sample $\bX$ of $n$ i.i.d. draws from $\cP$, and
  then having $\anl$ interact with $\mech(\bX)$ for $k$ rounds, where
  in each round $i$,
\begin{inparaenum}
\renewcommand{\labelenumi}{(\roman{enumi})}
\item $\anl$ selects $\phi_i$ (based on $a_1,\ldots,a_{i-1}$),
\item $\mech$ answers $a_i$.
\end{inparaenum}
The
\emph{(population) error} of $\mech$ is the random variable
  $$\err_\bX(\mech,\anl) = \max_i\abs{\phi_i(\cP)-a_i} \, .$$
  which depends on $\bX$ as well as the coins of $\mech$ and $\anl$. 

\begin{definition}\label{def:SQaccuracy}
  A query answering mechanism $\mathcal{M}$ is
  $(\alpha,\beta)$-accurate on i.i.d. data for $k$ queries if for every data analyst $\anl$
  and distribution $\mathcal{P}$, 
  we have
  $$\Pr\paren{\err_\bX(\mech,\anl)  \leq \alpha}\geq 1-\beta\,.$$ The probability is over the
  choice of the dataset $\bX\sim_{\text{i.i.d.}}\cP$ and the randomness of the mechanism and the
  analyst. Similarly, the \emph{expected error} of $\mech$ is the
  supremum, over distributions $\cP$ and data analysts $\anl$, of
$\Exp (\err_\bX(\mech,\anl))\,.$ We sometimes fix the distribution
$\cP$ and take the supremum only over analysts $\anl$. 
\end{definition}

It is important to note that this definition makes no assumptions on
how the analyst selects queries, except that the selection is based on
the outputs of $\mech$ and not directly on the data. The aim of this line of work is to design mechanisms with provable bounds on accuracy. We aim for mechanisms that are \emph{universal}, in the sense that they can be
used in any type of exploratory or adaptive workflow.

\subsection{Failures of Straightforward Approaches}
\label{sec:naive}

As mentioned above, there are a couple of natural approaches to this
problem. The first is to answer queries using each query's empirical
mean $\phi(\bx)$.  When queries are specified \emph{non}adaptively, a standard
argument shows that the population error of that strategy is
$\Theta\paren{\sqrt{(\log k) /n}}\, .$

In contrast, in the adaptive setting, the empirical mechanism's error
may be unbounded even with just two queries. 
For example, the query
$\phi$ may be selected such that  the low-order bits of $\phi_1(\bx)$
reveal all the entries of the data set $\bx$. In that case, the
analyst may construct a query $\phi_2$ which takes the value 1 for
values in the data set $\bx$, and 0 otherwise. The empirical mean
$\phi_2(\bx)$ will be 1, while the population mean $\phi_2(\cP)$
will be close to 0 for any distribution $\cP$ with sufficiently high entropy.

This last example seems contrived, since it requires seemingly
atypical structure from the initial query $\phi_1$. For example,
constraining the queries $\phi$ to be \emph{predicates} taking values
in $\{0,1\}$ seems to eliminate the problem. However, the example is
instructive for at least two reasons. First, it illustrates the role
that \emph{information about the data set} can play: learning $\bx$
allows the analyst to pose a query that is highly overfit to the
data set, and thus difficult for the
mechanism to answer accurately. Conversely, we will see that limiting
the information revealed about the data strongly limits overfitting.

Second, when the analyst asks more queries, one can construct much
more natural examples of analyses that go awry when 
using the empirical mean.  For instance,
consider a data set where each individual data point lies in
$\cX=\{0,1\}^{k-1}\times \{0,1\}$, where we think of the first $k-1$
bits as a vector of binary features, and the last bit as a
label. Consider a particular analyst (from~\cite{DFHPRR15science})  aiming to find a good classification rule
for the label. The analyst's first $k-1$ queries ask for the
success rate of each of the $k-1$ features in predicting the
label. In the $k$-th query, the analyst constructs a
classifier that takes a majority vote among those features that had
success rate greater than 50\%. On uniformly random data (where the
label is independent of the features), the mechanism will report the
success rate of this
last classifier to be $55\%$ when $k=\frac{n}{10}$, and $67\%$ when
$k=n$ (even though its success rate on the
population would be $1/2$). Generalizing the example somewhat, one
can show that even with very simple data distributions, the error of
empirical mechanism scales as $\Theta(\sqrt{k/n})$---exponentially larger
than the error one gets with nonadaptively specified
queries. Encapsulating this discussion, we have:
\as{Future: Add something about VC dimension and uniform convergence.}

\begin{prop}
  When answering $k$ nonadaptively specified queries, the empirical mechanism has expected error
  $\Theta\paren{\sqrt{\log(k) / n}}$. When answering $k$ adaptively
  selected queries, the empirical mechanism has expected error
  $\Omega \paren{\sqrt{k/n}}$, even for predicate queries on uniformly random data in $\{0,1\}^{k}$. 
\end{prop}

\paragraph{Data Splitting} Another natural
approach for handling adaptively specified queries is \emph{data
  splitting}: when $k$ is known in advance, one may divide the data
set into $k$ subsamples of $n/k$ points each, and answer the $i$-th
query using its empirical mean on the $i$-th data set. This approach
means that we can truly ignore adaptivity and use all the tools of
classical statistics; the downside is that we are limited to the
accuracy one can get with sample size $n/k$. The fact that we want a
bound that is uniform over all $k$ queries adds a further logarithmic
factor to the final error bound:

\begin{prop}
  When answering $k$ adaptively specified queries, the data splitting mechanism has expected error
  $\Theta(\sqrt{k(\log k) / n})$. 
\end{prop}

For both of these natural mechanisms, answering queries with error
$\alpha$, even with constant probability, requires $n$ to grow at
least as fast as $k/\alpha^2$. Can we do
better? How good a dependency on $\alpha$ and $k$ is possible?

\subsection{A Sample of Known Bounds}
\label{sec:known}

In fact, there are mechanisms that can answer a sequence of $k$
adaptively selected linear queries with much higher accuracy than that
provided by the straightforward approaches. Namely, for a given
accuracy $\alpha$, we can get mechanisms that work for $n$ that scales
only as $\sqrt{k}/\alpha^2$---a quadratic improvement in $k$. The
bounds below are stated in terms of expected error for simplicity; the
 underlying
arguments also provide high-probability bounds on the tail of this
error.

\begin{theorem}[\cite{DFHPRR15,BNSSSU16}]\label{thm:generalSQ}
There is a computationally efficient
mechanism for $k$
statistical queries  
with expected error $O\paren{\sqrt[4]{k}/\sqrt{n}}$.
\end{theorem}

A simple mechanism that achieves this bound is one that adds Gaussian
noise with standard deviation about $\sqrt[4]{k}/\sqrt{n}$ to each
query. 

One can give a different-looking mechanism---which we do not describe
in this survey---to automatically adjust to the actual ``amount'' of
adaptivity in a given sequence of queries. Specifically, imagine that
the $k$ queries are grouped into $r$ batches, where the queries in a
given batch depend on answers to queries in previous batches but not
on the answers to queries in the same batch. For example, in the
classification example of the previous section, the number of rounds
$r$ is only $2$.

\begin{theorem}[\cite{DFHPRR15}]\label{thm:SQ-rounds}
  If there are at most $r$ rounds of adaptivity, then there is a
  computationally efficient mechanism with expected error
  $O\paren{\sqrt{r (\log k) / n}}$.  The algorithm is \emph{not}
  given the partition of the queries into batches.
\end{theorem}

The ideas underlying the two previous algorithms can also be adapted
to give better results when we make further assumptions about the
class of allowed queries, or the universe from which the data are
drawn. One such result, due to \citet{DFHPRR15} (and tightened in
\cite{BNSSSU16}), recovers a logarithmic dependence on $k$ in exchange
for a dependence on the size of the universe $\cX$ in which the data
lie.

\begin{theorem}[\cite{DFHPRR15,BNSSSU16}]\label{thm:SQ-small-universes}
There is a
computationally inefficient mechanism
with expected error \\ $O\paren{\sqrt[6]{\log|\cX|}\sqrt[3]{(\log k)/n}}$. The
mechanism runs in time linear in $|\cX|$ (and not $\log|\cX|$ as one
would naturally want).
\end{theorem}

None of these upper bounds is known to be tight in all parameter regimes,
but some lower bounds are known, in particular showing that the
scaling $n=\Omega(\sqrt{k})$ cannot be improved, and that inefficiency
of the mechanism in Theorem~\ref{thm:SQ-small-universes} is necessary.

\begin{theorem}[\citet{HardtU14,SteinkeU15}]
  For every mechanism $\mech$ that answers $k$ adaptively selected
  linear queries, for a sufficiently large universe $\cX$ (with
  $\log|\cX|$ exponential in $n$), there exist
  a distribution $\cP$ and an analyst $\cA$ for which the mechanism's
  error $\err_{\bX}(\mech,\anl)$ is $\Omega\paren{\sqrt{k}/n}$ with
  constant probability. Furthermore, for mechanisms that answer
  faithfully with respect to both the distribution \emph{and} the data
  set (that is, they provide answers close to both $\phi_i(\bX)$ and
  $\phi_i(\cP)$), the bound can be strengthened to $\Omega\paren{\sqrt[4]{k}/\sqrt{n}}$

Finally, if we assume that one-way functions
  exist, then the bounds continue to hold when $\log|\cX|$ has polynomial size,   for
  polynomial-time mechanisms (but not for those that can take exponential
  time).  
\end{theorem}

We won't discuss the proof of these lower bounds here, but we note
that closing the gap between the upper and lower bounds remains an
intriguing open problem.

\subsection{Privacy and Distributional Stability}

The upper bounds above are obtained via a connection between adaptive
analysis and certain notions of algorithmic stability. Broadly,
algorithmic stability properties limit how much the output of an
algorithm can change when one of its inputs is changed. Different
notions of stability correspond, roughly, to different measures of
distance between outputs.  There is a long-standing connection between
algorithmic stability and expected generalization error (e.g.,
\citet{DevroyeW79,BousquetE02}). Essentially, \emph{stable algorithms
  cannot overfit}. It seems that if we could design
adaptive query-answering mechanisms that are stable in an appropriate sense, we could get
validity guarantees for adaptive data analysis.  

Alas, there is a hitch. Recall that our goal is to design mechanisms
that provide statistically valid answers no matter how the analyst
selects queries.  Even if each stage of the mechanism is stable, the
overall process might not be---in an adaptive setting, the analyst
ends up being part of the mechanism.

The resolution is to consider a \emph{distributional} notion of
stability. We will require that changing any single data point in
$\bx$ have a small effect on the distribution of the mechanism's
outputs. If we choose a distance measure on distributions that is
nonincreasing under postprocessing, then we can limit the effect of
the analyst's choices.

Specifically, we work with  ``differential privacy'', a notion of
stability introduced in the context of privacy of statistical
data. Differential privacy seeks to limit the information revealed
about any single individual in the data set. 

\begin{definition}[\cite{DMNS06,DworkKMMN06}]
An algorithm $\mech:\cX^n\rightarrow \mathcal{O}$ is
$(\eps,\delta)$-differentially private if for all pairs of
neighboring data sets $\bx, \bx' \in \cX^n$, and for all events $S
\subseteq \mathcal{O}$: $$\Pr[\mech(\bx) \in S] \leq
\exp(\eps)\Pr[\mech(\bx') \in S] + \delta\, .$$
\end{definition}

Differential privacy makes sense even for interactive mechanisms that
involve communication with an outside party: we simply think of the
outside party as part of the mechanism, and define the final output
of the mechanism to be the complete transcript of the communication between the
mechanism and the other party. 

Differential privacy is a useful design tool in the context of
adaptive data analysis because it is possible to  design interactive
differentially private algorithms \emph{modularly}, due to two related
properties: closure under postprocessing, and composition: 

\begin{prop}
  If $\mech:\cX^n\rightarrow \mathcal{O}$ is
  $(\eps,\delta)$-differentially private, and
  $f:\mathcal{O}\rightarrow \mathcal{O}'$ is an arbitrary (possibly
  randomized) mapping, then
  $f \circ \mech:\cX^n\rightarrow \mathcal{O}'$ is
  $(\eps,\delta)$-differentially private.
\end{prop}

\begin{prop}[Adaptive Composition~\cite{DworkRV10,KOV15}---informal]
 Let $\mech_1, \mech_2,...,\mech_k$ be a sequence of
 $(\eps,\delta)$-differentially private algorithms that are all run on
 the same data set, and selected
 adaptively (with the choice of $\mech_i$ depending on the outputs of
 $\mech_1,...,\mech_{i-1}$, but not directly on $\bx$). Then no matter how the adaptive selection
 is done, the resulting composed process is
 $(\eps',\delta')$-differentially private, for $\eps' \approx
 \eps\sqrt{k}$ and $\delta'\approx k\delta$.  
\end{prop}

Taken together, these two properties mean that in order to design
differentially private algorithms for answering linear queries, it is
sufficient to make sure the mechanism run at each stage is
differentially private. 

Perhaps even more importantly, in order to ensure statistical
validity---that is, accuracy with respect to the underlying
population---it suffices to design differentially private algorithms
that are accurate with respect to the sample $\bx$:

\begin{theorem}[Main Transfer Theorem~\cite{DFHPRR15,BNSSSU16}]
\label{thm:main-transfer}
Suppose a statistical estimator $\mech$  is 
$(\eps,\eps\cdot \delta)$-accurate with respect to its sample, that is, for
all data sets $\bx$,
$$\Pr(\max_i \left |a_i - \phi_i(\bx)\right | \leq \eps)\geq
1-\eps\cdot \delta\, .$$
If $\mech$ is also $(\eps,\eps\cdot\delta)$-differentially
private, then it is $(O(\eps),O(\delta))$-accurate with respect to the
population (Definition~\ref{def:SQaccuracy}).
\end{theorem}

This theorem underlies all two of the three upper bounds of the previous section
(Theorems~\ref{thm:generalSQ} and
\ref{thm:SQ-small-universes}). Each is derived by using existing
differentially private algorithms together with
Theorem~\ref{thm:main-transfer}. For
Theorem~\ref{thm:SQ-rounds}, \citet{DFHPRR15} used a different
argument, based on compressing the output of the algorithm to a small
set of possibilities; see Section~\ref{sec:info-theory}.

\subsection{A Two-stage Game, Stability  and ``Lifting''}
\label{sec:two-stage}

We conclude this section with an outline of the proof of the main
transfer theorem (Theorem~\ref{thm:main-transfer}). That theorem talks
about analyses with many stages of interaction, but it turns out
that the core of the argument lies in understanding a seemingly much
simpler, two-stage process.  

\begin{figure}
  \centering
  \includegraphics[height=1in]{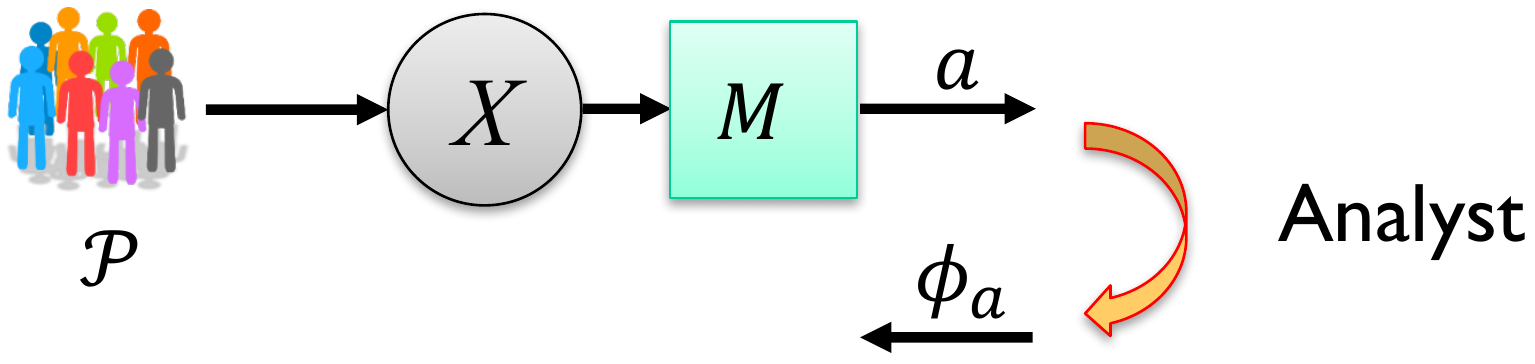}
  \caption{A two-stage overfitting game}
  \label{fig:twostage}
\end{figure}

Consider a two-stage setting in
which an analysis $\mech$ is run on data set $\bx$, and the analyst
selects a linear 
query $\phi:\cX\to [0,1]$ based on $\mech(\bx)$ (Figure~\ref{fig:twostage}). We say $\mech$
\emph{robustly $(\alpha,\beta)$-generalizes} if for all distributions $\cP$
over the domain $\cX$, for all strategies (functions) $\anl$ employed by the analyst, with probability at least $1-\beta$ over the
choice of $\bX\sim\cP^n$ and the coins of $\mech$, we have that
$|\phi(\bX)-\phi(\cP)|\leq \alpha$ where
$\phi=\anl(\mech(\bX))$. (Similarly, we may talk about the expected generalization error, that is, the maximum
over $\anl$ and $\cP$ of $\Exp(|\phi(\bX)-\phi(\cP)|)$.)

The quantification over all selection functions $\anl$ here is
critical---when the first phase of analysis satisfies the definition,
then a query asked in the following round cannot overfit to the data
(except with low probability), no matter how it is selected.

Differential privacy (and a few other distributional notions of
stability, such as KL-stability~\cite{BNSSSU16,WangLF16klprivacy})
limits the adversary's score in this game. This connection had been
understood for some time---for example, McSherry observed that it
could be used to break up dependencies in a clustering algorithm, and
\citet{BassilyST14} used a weak version of the connection to bound the
population risk of differentially private empirical risk
minimization. 

However, the application to adaptive data analysis--and especially the
understanding of the importance of post-processing to the design of
universal mechanisms---came recently, in \cite{DFHPRR15}. Their
initial result was subsequently sharpened, to obtain the following
tight connection:

\begin{theorem}[Differentially Private Algorithms Cannot Overfit \cite{BNSSSU16}]
  If $\mech$ is $(\eps, \eps \delta)$-differentially private,
  then it is $(O(\eps), O(\delta))$-robustly generalizing.  
\end{theorem}

\paragraph{Lifting to Many Stages}
Bounds
on the two-stage game can be ``lifted'' to provide bounds on the
$k$-phase game either through a sequential application of the bound to
each round \cite{DFHPRR15} or through a more holistic argument, called
the \emph{monitor technique}~\cite{BNSSSU16}, that yields
Theorem~\ref{thm:main-transfer} (and Theorems~\ref{thm:generalSQ}
and 
\ref{thm:SQ-small-universes}). 

The monitor argument is a thought experiment---we argue that for any
multi-stage process, there is a two-stage process in which the error
on the population 
equals the maximum  population error over all $k$ stages of the
original process. The argument 
applies quite generally, but it is a bit simpler
to explain under the assumption that the mechanism answers queries
accurately with respect to the data set $\bX$. The idea, given an
interaction between an analyst and adaptively selected mechanisms $\mech_1, \mech_2,
...,\mech_k$, is to encapsulate the analyst and mechanisms into a
single fictional entity $\mech$ which gets, as additional input, the
underlying distribution $\cP$. The fictional $\mech$ executes an interaction  and
then outputs a single query $\phi^*$---the one which maximizes the
population error over all stages $i$. 

\begin{figure}
  \centering
  \includegraphics[height=2in]{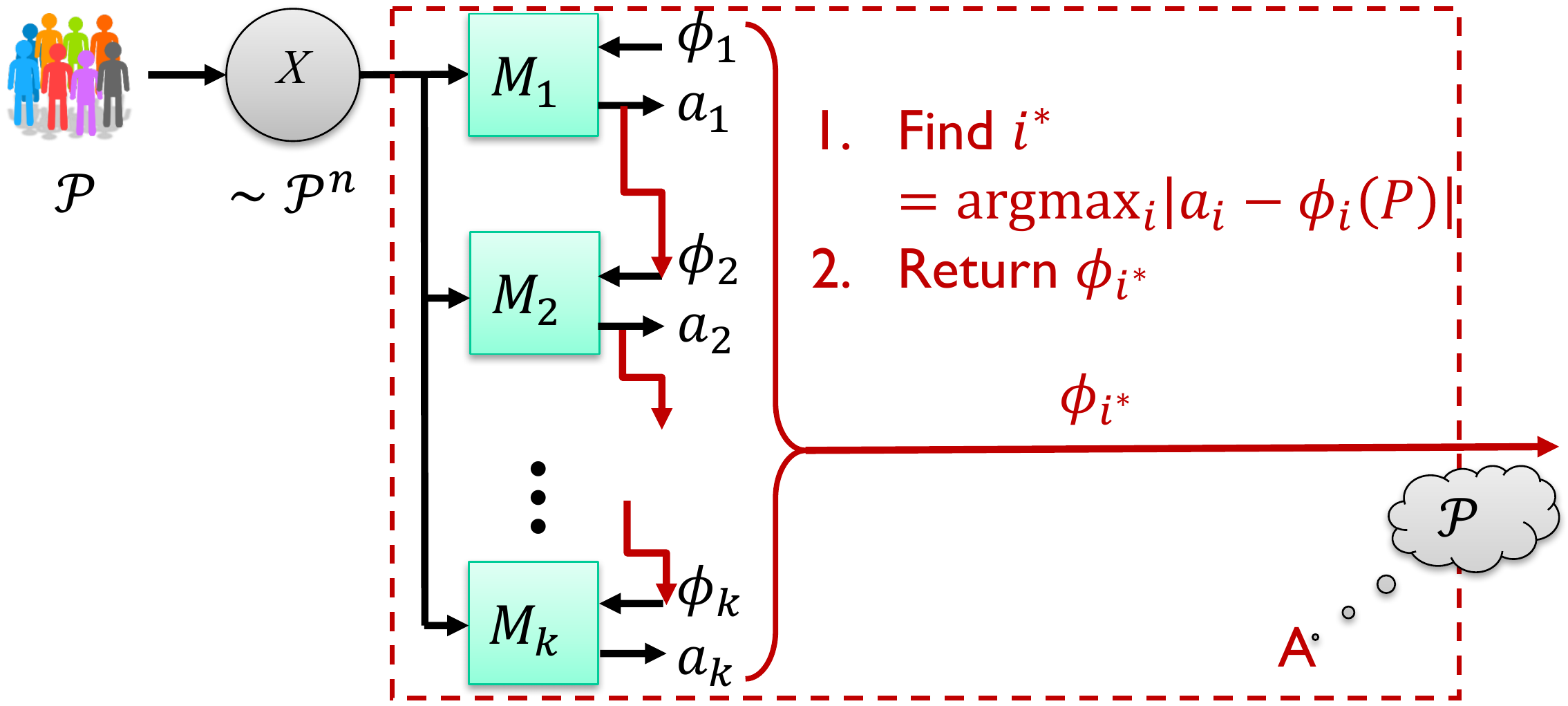}
  \caption{The ``monitor'' argument for lifting results about a
    two-stage game to many stages}
  \label{fig:monitor}
\end{figure}

\paragraph{Beyond linear queries} The techniques described in this
section extend to problems that are not described by estimating the
mean of a bounded linear functional. One important class is minimizing a
decomposable loss function where each individual contributes a bounded
term to the loss function~\cite{BNSSSU16}.

\paragraph{A re-usable holdout} The techniques described in this
section can appear somewhat onerous for the analyst, since they
require accessing data via differentially private algorithms. As
pointed out by \citet{DFHPRR15science}, however, one need not limit
access to the entire data set in this way. In fact, a more pragmatic
approach is to give most of the data ``in the clear'' to the analyst,
and protect only a small holdout set via the techniques discussed
here. This still allows one to \emph{verify} conclusions soundly, but
additionally allows full exploratory analysis, as well as repeated
verification (``holdout re-use'').

\section{The Intrigue of Information Measures}
\label{sec:info-theory}

Despite the generality of the approach of the previous section, many
important classes of analyses are not obviously amenable to those
techniques; in particular, problems that are not easily stated in
terms of a numerical estimation task. 

Consider the problem of hypothesis testing. Crudely, given
a set of distributions $\cH$ (called the null hypothesis), we ask if
the data set is ``unlikely to have been generated'' by a distribution
$P \in \cH$. More precisely, we select an event $T$ (the \emph{acceptance
region}) such that
$\Pr_{\bX \sim \cP^n}(\bX\in T)$ is at most a threshold $\sig$ (often 0.05) for all
distributions in $\cH$. If it happens that the observed data $\bX$ lie
outside of $T$, the null hypothesis is said to be rejected. If this
happens when the true distribution $\cP$ is actually in $\cH$, then we
say a \emph{false discovery} occurs. Hypothesis tests play a central role
in modern empirical science (for better or for
worse), and techniques to control false discovery
in the classic, nonadaptive setting
are the focus of intense study. \as{Future: add citations.} Despite this, very little is known
about hypothesis tests in adaptive settings.  

Adaptivity arises when the event $T$ is selected based on earlier
analysis of the same data -- conditioned on those earlier results $Y=\mech(\bX)$,
the probability that the test rejects the null hpyothesis given $\bX$
might be much higher than $\sig$ even if $\cP$ lies in $\cH$. 

How much higher it can be depends on $\mech$ and---as we will see---on
several measures of the information leaked by $\mech$. To formalize
this, consider a game similar to the overfitting game, in which the
analyst $A$, given $Y=\mech(\bX)$, selects an arbitrary event $T_Y = A(Y)$ (which
depends on $Y$). For a particular output $y$ of $\mech$, the analyst's ``score'' is 
$$\score_{\cP,\mech,A,\sig}(y) =
\begin{cases}
  \Pr_{\bX \sim \cP^n} (\bX\in T_y | Y=y) &  \text{if }\Pr_{\bX \sim \cP^n} (\bX\in
  T_y) \leq \sig \, ,\\
  0 & \text{if }\Pr_{\bX \sim \cP^n} (\bX\in T_y) > \sig \, .
\end{cases}
$$

Now consider the analyst's \emph{expected} score in this game:
$\displaystyle \eta_{\cP,\mech,A}(\sig) = \Exp_{Y}
\big(\score_{\cP,\mech,A}(y)
\big)  
$.
As we will see below, the analyst's score in this game can be bounded
using various definitions of the \emph{information} leaked about $\bX$ by $Y$. This
score also plays a key role in controlling false discovery:

\begin{prop}Bounding the score $\eta$ has several important implications: 
  \begin{compactenum}\item 
\textbf{(False discovery~\cite{RRST16})} \footnote{The
    definitions here differ somewhat from those of~\cite{RRST16}; in
    particular, our function $\eta$ is the inverse of a ``$p$-value
    correction function'' from ~\cite{RRST16}.}
  If $Y=\mech(\bX)$ is used by $A$ to select a hypothesis test with
  significance $\sig$, then the probability of false discovery is at
  most $\eta_{\cP,\mech,A}(\sig)$.
\item \textbf{(Robust generalization~\citep{DFHPRR15nips})} If $Y=\mech(\bX)$ is used by $A$
  to select a bounded linear query $\phi_Y$, then
  $\Pr\bparen{\left|\phi_Y(\bX) - \phi_Y(\cP)\right| \geq t/\sqrt{n}}\leq
  \eta_{\cP,\mech,A}\left(e^{-t^2/3}\right)$.
\end{compactenum}
\end{prop}


We are interested in universal bounds that hold no matter how the
analyst uses the output $Y$, and no matter the original input
distribution. To this end, we define 
$$\eta_\mech(\sig) = \sup_{\cP,A} \left(\eta_{\cP,\mech,A}(\sig)\right)\, .$$

\subsection{Information and Conditioning}

The function $\eta_M$ measures how much probabilities less than $\sig$
can be amplified by conditioning on $Y$, on average over values of $Y$.

For several one-shot notions of mutual information, we have that if a
procedure leaks $k$ ``bits'' of information we have
$\eta_{\cP,\mech,A}(\sig)  \approx \sig\cdot 2^k$. Unfortunately, such a clean
relationship is not known for the standard notion of Shannon mutual
information.  Instead, we consider two other notions here. 

Fix two random
variables $X,Y$ with joint distribution given by $p_{X Y}(x,y) =
\Pr(X=x,Y=y)$ and marginals $p_X(\cdot)$ and $p_Y(\cdot)$. Consider
the information loss $$I_{x,y}= \log
\left(\frac{p_{X Y}(x,y)}{p_X(x)p_Y(y)}\right)\, .$$

The standard notion of mutual information is the expectation of this
variable: $I(X;Y) = \Exp\left( I_{X ,Y}\right)$. 
The \emph{max-information} \cite{CiganovicBR14,DFHPRR15nips} between $X$ and $Y$ is
the supremum of this variable. Unfortunately, for many interesting
procedures, the max-information is either unbounded or much larger
than the mutual information. 

One can get a more flexible notion by instead considering a
high-probability bound on the information loss: we say the
\emph{$\beta$-approximate max information} between $X$ and $Y$ is at
most $k$ (written $I_{\infty}^\beta(X;Y)\leq k$) if
$\Pr_{(x,y)\sim(X,Y)}(I_{x,y}\leq k)\geq 1-\beta$.~\footnote{This
  condition is not exactly the definition of max-information of
  \cite{DFHPRR15nips} (which requires that for all events $E$,
  $\Pr((X,Y)\in E) \leq 2^k\Pr((X',Y)\in E)$ where $X'$ is idnetically
  distributed to $X$ but independent from $Y$). The definition her
  implies that of \cite{DFHPRR15}.}~\footnote{The
  $\beta$-approximate max information is equivalent to a
  \emph{smoothed} version of max-information
  \cite{RenWol04a,RennerW05,TomamichelCR10,VitanovDTR13,CiganovicBR14}, in which
  we ask that the pair $X,Y$ be within statistical distance $\beta$ of
  a joint distribution with $I_\infty^0(X,Y)\leq k$. See Corollary~8.7
in \citet{BunS16} for details.}

For many algorithms of interest, the approximate max-information turns
out to be very close to the mutual information but, by providing a
bound on the upper tail of $I_{X ,Y}$, allows for more precise control
of small-probability events.

We can also define a related quantity, which we call the
\emph{expected log-distortion}:
$$L_\infty (X;Y) = \log \Exp_{y\sim Y}\left( \sup_{x}\left( 2^{I_{x,y}}
\right) \right)
= \log \Exp_{y\sim Y}\left( \sup_{x} \frac{p_{X Y}(x,y)}{p_X(x)p_Y(y)} \right)
\, .
$$
This notion of information leakage is not symmetric in $X,Y$. It is
closely related to, but in general different from, the min-entropy
leakage $H_\infty(Y|X) - H_\infty(X)$ ~\cite{DORS08,AlvimCPS12,EspinozaS13,AlvimCMMPS14,AlvimCMMPS16}. \as{Future:
  add CDP-style bounds.}

Of these notions, expected log distortion is the strongest since it
upper bounds the other two: 
$I(X;Y)\leq L_\infty (X;Y)$ and
$I_\infty^{\beta}(X;Y)\leq L_\infty (X;Y) + \log(1/\beta)$.

\begin{theorem} \label{thm:info-gen}
For every mechanism $\mech$ and dsitribution $\cP$ s.t. $\bX\sim\cP$:
  \begin{compactenum}
  \item If $I_\infty^\beta (\bX;Y) \leq k$, then for every analyst $A$,
    $\eta_{\cP,\mech,A}(\sig)\leq \sig\cdot 2^k + \beta$, and \\
    $\mech$ robustly $(\alpha,2\beta)$-generalizes for
    $\alpha=\sqrt{(k+2\ln(1/\beta))/n}$ on $\cP$. 
  \item If $L_\infty (\bX;\mech(\bX)) \leq k$,
    then for every analyst $A$,
    $\eta_{\cP,\mech,A}(\sig)\leq \sig \cdot 2^k$, and \\
    for every $\beta>0$, $\mech$ robustly $(\alpha,\beta)$-generalizes for
    $\alpha=\sqrt{(k+2\ln(1/\beta))/n}$ on $\cP$. 
  \end{compactenum}
\end{theorem}

We know much weaker implications based only on bounding the mutual
information. Most significantly, the  bounds for general hypothesis
testing are exponentially weaker than those one gets from the one-shot
measures above.

\begin{prop}[\cite{russo2015controlling,RRST16}]
  If $I (\bX;\mech(\bX)) \leq k$,
  then 
  \begin{inparaenum}\renewcommand{\labelenumi}{(\roman{enumi})}
  \item 
    for every analyst $A$,
    $\eta_{\cP,\mech,A}(\sig)\leq \frac{k+1}{\log_2(1/\sig)}$,
    and 
  \item for
    every $\beta>0$, $\mech$ robustly $(\alpha,\beta)$-generalizes for
    $\alpha=O(\sqrt{k/n\beta})$ on $\cP$.
  \end{inparaenum}
\end{prop}





\subsection{What procedures have bounded one-shot information
  measures?}

The information-theoretic framework of the previous subsection
captures several other classes of algorithms that satisfy robust
generalization guarantees. In addition to unifying the previous work,
this approach shows that these classes of algorithms allow for
principled post-selection hypothesis testing.

The most important of these, currently, is for the class of
differentially private algorithms: 

\begin{theorem}[Informal, see \cite{RRST16}]\label{thm:DPmaxinfo}
If $\mech$ is $(\eps,\delta)$-differentially private, and the entries of $\bX$ are independent, then $I_\infty^\beta(\bX;\mech(\bX)) =O(\eps^2 n)$ for $\beta = O(n\sqrt{\delta/\eps})$.
\end{theorem}

This result, together with Theorem~\ref{thm:info-gen}, implies that
differentially-private algorithms are $(O(\eps), O(n\sqrt{\delta/\eps})$-robustly generalizing
for data drawn i.i.d from any distribution $\cP$. It essentially
recovers the results of the previous section on linear queries (with a
worse value of $\beta$), but additionally applies to more general
problems such as hypothesis testing.

\paragraph{Description length~\cite{DFHPRR15nips}} In many cases, the outcome of a
statistical analysis can be compressed to relatively few bits---for
example, when the outcome is a small set of selected features. If the
output of $\mech$ can be compressed to $k$ bits, then the expected
log-distortion $L_\infty(\bX;\mech(\bX))$ is at most $k$ bits. An
argument along these lines was used implicitly in \cite{DFHPRR15} to
prove Theorem~\ref{thm:SQ-rounds}.
~
\as{Future: explain rounds of adaptivity}

\paragraph{Compression Schemes} Another important class of statistical
analyses that have good (and robust) generalization properties are
compression learners~\cite{LW86}. These process a data set of $n$ points to obtain
a carefully selected subset of only $k<<n$ points, and finally produce
an output fit to those points. A classic example is support vector
machines: in $d$ dimensions, the final classifier is determined by
just $d+1$ points in the data set.

\citet{CLNRW16} used classic generalization results for such learners
to show that they satisfy robust generalization guarantees. The
classic results as well as those of \citet{CLNRW16} 
can be rederived from the following lemma (new, as far as we
know) bounding the
information leaked by a compression scheme about those points that are
\emph{not} output by the scheme.

\begin{lemma}[Compression schemes] \label{thm:compressionInfo}
Let $\mech:\cX^n\to \cX^k$ be any algorithm that takes a data set $\bx$
of $n$ points and 
outputs a subset $\bx_{out}=\mech(\bx)$ of $k$ points from $\bx$. Let
$\bx_{in}\in \cX^{n-k}$ denote the remaining data points, so that
$\bx_{in} \cup \bx_{out} = \bx$ (as multisets). For any distribution $\cP$ on $\cX$, if $\bX\sim\cP^n$,
then $$L_\infty(\bX_{in} ; \bX_{out})\leq \log_2\binom n k\, .$$
\end{lemma}

\citet{CLNRW16}  used the robust generalization properties of
compression learners to give robustly generalizing algorithms for
learning any PAC-learnable concept class. In particular, this implies
robustly generalizing algorithms for tasks that do not have
differentially private algorithms, such as learning a threshold
classifier with data from the real line.

\subsection*{Notes and Acknowledgments}
\label{sec:model-discuss}

I am grateful to many colleagues for introducing me to this topic and
insightful discussions. Among others, my thanks go to Raef Bassily,
Cynthia Dwork, Moritz Hardt, Kobbi Nissim, Ryan Rogers, Aaron Roth,
Thomas Steinke, Uri Stemmer, Om Thakkar, Jon Ullman, Yu-Xiang Wang.
Paul Medvedev provided helpful comments on the writing, making the
survey a tiny bit more accessible to nonexperts.  Finally, I thank
Michael Langberg, this newsletter's editor, for soliciting this
article and graciously tolerating delays in its preparation.

Although I tried to cover the main ideas in a recent line of work,
that line is now diverse enough that I could not encapsulate
everything here. Notable omissions include the ``jointly Gaussian''
models of \citet{russo2015controlling} and \citet{WangLF16}, which assume that the
analyst is selecting among a family of statistics that are jointly
normally distributed under $\cP$ , work of \citet{Elder16a,Elder16b} on a Bayesian framework that
encodes a further restriction that the mechanism ``know as much'' as
the analyst about the underlying distribution $\cP$, and the ``typcial
stability'' framework~\cite{BF16,CLNRW16}. There are no
doubt other contributions that I lost in my effort to consolidate. 
Some ideas in this survey, notably the information-theoretic viewpoint
on compression-based learning, have not appeared elsewhere; they
have not been peer-reviewed.

\begin{small}
  \bibliographystyle{abbrvnat} \bibliography{refs}
\end{small}

\end{document}